\pdfoutput=1

\RequirePackage{tikz} % for some reason this has to go before the documentclass declaration

\documentclass[letterpaper, 10 pt, conference]{ieeeconf}  % Comment this line out if you need a4paper

\IEEEoverridecommandlockouts                              % This command is only needed if 
\usepackage{comment}
\usepackage{amsmath,amssymb,amsfonts}
\usepackage{algorithm}
\usepackage[noend]{algpseudocode}
\usepackage{graphicx}
\usepackage{textcomp}
\usepackage{booktabs}
\usepackage{hyperref}
\usepackage{xspace}
\usepackage{balance}
\usepackage{array}
\usetikzlibrary{arrows,calc,positioning,shapes,fit}
\usepackage[sorting=none]{biblatex}
\usepackage{caption}
\let\labelindent\relax
\usepackage{enumitem}
\usepackage{array}
\usepackage{multirow}

\tikzstyle{block} = [rectangle, draw, text centered, rounded corners, minimum height=2em]
\tikzstyle{eq} = [circle, draw, minimum height=1.7em, inner sep=1pt] 
\makeatletter
\newcommand{\gettikzxy}[3]{%
  \tikz@scan@one@point\pgfutil@firstofone#1\relax
  \edef#2{\the\pgf@x}%
  \edef#3{\the\pgf@y}%
}
\makeatother

\usepackage{biblatex}

\def\BibTeX{{\rm B\kern-.05em{\sc i\kern-.025em b}\kern-.08em
    T\kern-.1667em\lower.7ex\hbox{E}\kern-.125emX}}
\title{A Comparative Study of SMT and MILP for the Nurse Rostering Problem \\
\author{
    Alvin Combrink$^{1}$, Stephie Do$^1$, Kristofer Bengtsson$^2$, Sabino Francesco Roselli$^1$, Martin Fabian$^1$% <-this % stops a space
    \thanks{$^{1}$Division of Systems and Control, Department of Electrical Engineering, Chalmers University of Technology, G{\"o}teborg, Sweden
        {\tt\footnotesize \{combrink, rsabino, fabian\}@chalmers.se}}
    \thanks{$^{2}$Research and Technology Development, Group Trucks Operations, Volvo AB G{\"o}teborg, Sweden 
        {\tt\footnotesize \{kristofer.bengtsson@volvo.se\}}}
}
\thanks{This work was supported by the Wallenberg AI, Autonomous Systems and Software Program (WASP) funded by the Knut and Alice Wallenberg Foundation.}
}

\newcommand{\GitHubLink}{{\scriptsize\url{https://github.com/Adcombrink/NRP-GenericConstraints}}\xspace}

\newcommand{\Staff}{\mathcal{P}}
\newcommand{\numstaff}{n_{\person[]}}
\newcommand{\stdperson}{j}
\newcommand{\person}[1][\stdperson]{p_{#1}}
\newcommand{\personq}[1][\stdperson]{Q_{#1}}
\newcommand{\persondwl}[1][\stdperson]{\mathit{dw}_{#1}}

\newcommand{\Qualifications}{\mathcal{Q}}

\newcommand{\Shifts}{\mathcal{S}}
\newcommand{\numshifts}{n_{\shift[]}}
\newcommand{\stdshift}{i}
\newcommand{\shift}[1][\stdshift]{s_{#1}}
\newcommand{\shifttype}[1][\stdshift]{t_{#1}}
\newcommand{\shiftsd}[1][\stdshift]{\mathit{sd}_{#1}}
\newcommand{\shiftdur}[1][\stdshift]{\mathit{d}_{#1}}

\newcommand{\shiftwl}[1][\stdshift]{\mathit{w}_{#1}}
\newcommand{\shiftqr}[1][\stdshift]{\mathit{RQ}_{#1}}
\newcommand{\shiftqp}[1][\stdshift]{\mathit{QP}_{#1}}

\newcommand{\Startdays}{\mathcal{D}}

\newcommand{\ShiftFiltersd}[1]{\Shifts^{sd}(#1)}

\newcommand{\ShiftFiltertype}[2]{\Shifts^{t}(#2, #1)}

\newcommand{\Overlappings}{\mathcal{O}}
\newcommand{\numoverlap}{n_{\overlap[]}}
\newcommand{\stdoverlap}{k}
\newcommand{\overlap}[1][\stdoverlap]{o_{#1}}
\newcommand{\overlapstaff}[1][\stdoverlap]{{\Staff}_{#1}}
\newcommand{\overlapshifts}[1][\stdoverlap]{{\Shifts}_{#1}}

\newcommand{\GCparamStaff}[1][]{{\color{orange} \widetilde\Staff#1}}

\newcommand{\GCparamShifts}[1][]{{\color{orange} \widetilde\Shifts#1}}

\newcommand{\GCparamxbare}{{\color{blue} x}}
\newcommand{\GCparamybare}{{\color{blue} y}}
\newcommand{\GCparamnbare}{{\color{blue} n}}
\newcommand{\GCparammbare}{{\color{blue} m}}
\newcommand{\GCparamubare}{{\color{blue} u}}
\newcommand{\GCparamvbare}{{\color{blue} v}}
\newcommand{\GCparamx}{\GCparamxbare}
\newcommand{\GCparamy}{\GCparamybare}
\newcommand{\GCparamn}{\GCparamnbare}
\newcommand{\GCparamm}{\GCparammbare}
\newcommand{\GCparamu}{\GCparamubare}
\newcommand{\GCparamv}{\GCparamvbare}

\newcommand{\GCvard}{{\color{purple} d}}

\newcommand{\assignment}[1][\stdperson, \stdshift]{a_{#1}}

\newcommand{\GCvarInter}[1][]{{\color{violet} \delta#1}}
\newcommand{\GCconstBigM}[1][]{{\color{violet} M#1}}

\newcommand{\boolAND}[2]{\bigwedge_{#1}^{#2}}
\newcommand{\boolOR}[2]{\bigvee_{#1}^{#2}}
\newcommand{\boolsum}[1]{\underset{#1}{\bigoplus}}
\newcommand{\boolmap}{\oplus}
\newcommand{\twodots}{\mathinner {\ldotp \ldotp}}
\newcommand{\IntRange}[2]{ [ #1 \twodots #2 ] }

\newcommand{\ProblemShiftName}[3]{#1\,#2#3}
\newcommand{\ProblemShiftNameNurseD}[1]{\ProblemShiftName{Nurse}{D}{#1}}
\newcommand{\ProblemShiftNameNurseE}[1]{\ProblemShiftName{Nurse}{E}{#1}}
\newcommand{\ProblemShiftNameNurseN}[1]{\ProblemShiftName{Nurse}{N}{#1}}
\newcommand{\ProblemShiftNameDoctorD}[1]{\ProblemShiftName{Doctor}{D}{#1}}
\newcommand{\ProblemShiftNameDoctorE}[1]{\ProblemShiftName{Doctor}{E}{#1}}
\newcommand{\ProblemShiftNameDoctorN}[1]{\ProblemShiftName{Doctor}{N}{#1}}
\newcommand{\ProblemShiftNameDoctorS}[1]{\ProblemShiftName{Doctor}{S}{#1}}
\newcommand{\ProblemShiftNameCNurseD}[1]{\ProblemShiftName{C.Nurse}{D}{#1}}
\newcommand{\ProblemShiftNameCNurseE}[1]{\ProblemShiftName{C.Nurse}{E}{#1}}
\newcommand{\ProblemShiftNameAdmin}[1]{\ProblemShiftName{Admin}{}{#1}}

\begin{document}
\maketitle
\thispagestyle{empty}
\pagestyle{empty} 
%%%%%%%%%%%%%%%%%%%%%%%%%%%%%%%%%%%%%%%%%%%%%%%%%%%%%%%%%%%%%%
\begin{abstract}
The effects of personnel scheduling on the quality of care and working conditions for healthcare personnel have been thoroughly documented. 
However, the ever-present demand and large variation of constraints make healthcare scheduling particularly challenging.
This problem has been studied for decades, with limited research aimed at applying Satisfiability Modulo Theories (SMT). SMT has gained momentum within the formal verification community in the last decades, leading to the advancement of SMT solvers that have been shown to outperform standard mathematical programming techniques.

In this work, we propose generic constraint formulations that can model a wide range of real-world scheduling constraints. Then, the generic constraints are formulated as SMT and MILP problems and used to compare the respective state-of-the-art solvers, Z3 and Gurobi, on academic and real-world inspired rostering problems.
Experimental results show how each solver excels for certain types of problems; the MILP solver generally performs better when the problem is highly constrained or infeasible, while the SMT solver performs better otherwise. 
On real-world inspired problems containing a more varied set of shifts and personnel, the SMT solver excels.
Additionally, it was noted during experimentation that the SMT solver was more sensitive to the way the generic constraints were formulated, requiring careful consideration and experimentation to achieve better performance.
We conclude that SMT-based methods present a promising avenue for future research within the domain of personnel scheduling. 
\end{abstract}
%%%%%%%%%%%%%%%%%%%%%%%%%%%%%%%%%%%%%%%%%%%%%%%%%%%%%%%%%%%%%%
\section{INTRODUCTION}
\label{sec:Introduction}
Personnel scheduling, or \emph{rostering}, is crucial in healthcare.
Services depend heavily on human resources scheduled in shifts to meet highly variable and continuous demands.
This requires 24/7 coverage while accounting for staff fatigue, workload balancing, matching the many staff qualifications with shift requirements, and so on. 
These stringent conditions make healthcare scheduling particularly challenging.
%Rising global healthcare demand has outpaced supply in many regions~\cite{who2008workforce}, highlighting the need to improve efficiency and workplace conditions. 
Evidence shows that effective rostering has an impact on efficiency and workplace conditions, as poor scheduling can impair alertness and cognitive performance~\cite{ganesan2019impact, rajaratnam2013sleep}, and increase risks of self-injury or commuting accidents~\cite{rajaratnam2013sleep, barger2009neurobehavioral}. However, predictable schedules that consider staff needs can mitigate these risks~\cite{williams2022stable, wilson2002impact}.

There exists a myriad of variants of the \emph{Nurse Rostering Problem} (NRP)~\cite{cheang2003nurse} and \emph{Physician Scheduling Problem} (PSP)~\cite{erhard2018state}, with wide-ranging variations in the number of shifts, personnel, and constraints.
%The most prominent healthcare rostering problems are the nurse rostering problem (NRP)~\cite{cheang2003nurse} and the physician scheduling problem (PSP)~\cite{erhard2018state}, for which there exist a myriad of variations.
For example, \cite{santos2016integer} uses the common three shifts per day (Morning, Evening and Night), \cite{burke2008hybrid} extends this to four (Day, Early, Late and Night) while \cite{burke2001memetic} includes up to $15$ duties per day. 
Comprehensive accounts of rostering variants and methods can be found in e.g. \cite{erhard2018state, burke2004state, ernst2004staff, van2013personnel}.
The majority of the rostering approaches reviewed in~\cite{erhard2018state} use mathematical programming techniques. 
However, there is an apparent shift toward combining exact methods with heuristics to leverage performance gains at the cost of optimality guarantees. It is therefore interesting to evaluate if performance gains can be found over standard mathematical programming techniques without losing such guarantees.

Satisfiability Modulo Theories (SMT) has gained momentum within the formal verification community in the last two decades, driving advancements in SMT solvers such as Z3~\cite{de2008z3}, MathSAT5~\cite{cimatti2013mathsat5}, and OpenSMT~\cite{bruttomesso2010opensmt}. 
By extending propositional satisfiability to incorporate decision procedure theories involving predicate logic and quantifiers, SMT benefits from increased expressibility while retaining some of the performance of SAT solvers.
The viability of SMT solvers has been demonstrated.
Notably, Z3 yielded competitive results~\cite{roselli2018smt} compared to the commercial Mixed Integer Linear Program (MILP) solver Gurobi~\cite{gurobi} on the job-shop scheduling problem.
Similar results were shown on the rotating workforce problem when comparing SMT with two techniques based on network-modeling and constraint programming~\cite{erkinger2017personnel}.
\emph{Weighted SMT} was used to address the NRP with both hard and soft constraints by weighting constraints in the objective function~\cite{ansotegui2013solving}. However, no comparisons were made with other methods.

Of the research-developed NRP methods, only around $30$\% are implemented in practice~\cite{kellogg2007nurse}. A likely explanation for this is the large variability in the types of constraints that a real-world healthcare department places on its rosters, and the difficulty to model these constraints. Many methods build upon too simplified models to be applicable to real-world problems~\cite{burke2004state}.
Formulating \emph{generic constraints} is one way to address this.
Perhaps one of the earliest works implementing generic constraints is presented in~\cite{okada1992approach}.
More recently, generic constraints for the PSP where formulated with both hard and soft constraints~\cite{rousseau2002general}. It was shown that even a relatively few generic constraints could model a moderate range of requirements by applying their model to two different wards.
%More recently, the PSP was addressed using generic constraints that could then model a specific hospital ward's set of constraints~\cite{rousseau2002general}. Both hard and soft constraints where included. Even a relatively few generic constraints could model a moderate range of requirements, making it possible to model two different wards. 
% This shows that by designing constraints to be generic in nature and thereafter tailored to specific scheduling problems, proposed methods can gain a larger degree of sustainable, practical usefulness.

The contributions of this paper build upon the work in \cite{combrink_do_2021} in cooperation with the Queen Silvia Children's Hospital in Gothenburg, Sweden.
First, we propose a set of generic constraints that can model a wide range of real-world scheduling constraints.
Second, we formulate the generic constraints as both SMT and MILP problems, and 
%%MF perform comparisons of 
compare
the respective state-of-the-art solvers, Z3 and Gurobi.
%%MF Maybe a very short summary of the results?
Experimental results highlight strengths for each solver; Gurobi excels on highly constrained instances near or beyond the border of infeasibility, while Z3 performs better on less constrained instances. In conclusion, which solver to use depends on the underlying problem.

%%MF The article outline is as follows: ... 

\section{Problem Definition}
\label{sec:ProblemDefinition}

The goal of scheduling in this context is to find an assignment of \emph{staff members} to \emph{shifts} while fulfilling all constraints. Only one staff member can be assigned to a shift.

Let the \emph{personnel set} $\Staff = \IntRange{1}{\numstaff}$ contain the indexes corresponding to each staff member. 
A staff member $\stdperson\in\Staff$ has a \emph{desired workload} $\persondwl\in\Re$,
which is the desired amount of workload that they are assigned during the scheduling period.
Additionally, a staff member has a set of \emph{qualifications} $\personq \subseteq \Qualifications$, where $\Qualifications$ is the set of all qualifications, e.g. ``Administrator'' or ``Nurse''.
% that staff members may have and shifts may require. 
% The desired workload represents how much a staff member desires to work during the scheduling period. This is typically measured as a percentage of a typical amount of work per week.

Similarly, let the \emph{shift set} $\Shifts = \IntRange{1}{\numshifts}$ contain indexes for each shift. 
A shift $\stdshift\in\Shifts$ has a \emph{shift type} $\shifttype$ representing a category that the shift is included in, e.g. ``Day Shift'' or ``Anesthesiologists Shift''; 
\emph{start day} $\shiftsd \in \Startdays$ where $\Startdays=\IntRange{1}{T}$ contains an index for each day of the $T$-day long scheduling period;
a \emph{duration} $\shiftdur\in\Re$;
a \emph{workload} $\shiftwl\in\Re$ representing how burdensome a shift is, which is not necessarily equal to the duration;
and a set of \emph{required qualifications} $\shiftqr \subseteq \Qualifications$, where staff member $\stdperson\in\Staff$ is qualified for shift $\stdshift\in\Shifts$ if $\shiftqr \subseteq \personq$.
% The workload $\shiftwl$ represents how burdensome it is to work the shift. In many cases, this is equal to the duration, however, some duties are more burdensome than others and therefore they are not always equal. 
% Hence, the workload is separate from the duration of a shift.
Let
\begin{align*}
    \shiftqp = \left\{ \stdperson \in \Staff \mid \shiftqr \subseteq \personq \right\}
\end{align*}
be the set of \emph{qualified personnel} for shift $\stdshift\in\Shifts$.

In many real-world hospital wards, it is permissible for certain shifts which overlap in time to be assigned to the same staff member. This occurs in cases where, for example, the tasks of the shifts are compatible so that one person can handle both, but they are scheduled separately due to qualification requirements or staffing flexibility.
We refer to these as \emph{overlapping shifts}.
For $n$ shifts, there will be $2^n - n - 1$ number of overlapping combinations. Each combination is designated an index in the \emph{overlapping combinations} set $\Overlappings = \IntRange{1}{\numoverlap}$.
An \emph{overlapping combination} $\stdoverlap \in \Overlappings$ corresponds to 
a set of overlapping shifts $\overlapshifts \subseteq \Shifts$ and a set of personnel $\overlapstaff$.
Every person in $\overlapstaff$ must be qualified for every shift in $\overlapshifts$, i.e. $\overlapstaff \subseteq \bigcap_{\stdshift\in\overlapshifts} \shiftqp$, and are \emph{allowed} to be assigned to all these shifts at once.
We distinguish between being \emph{allowed} and being \emph{assigned} to $\overlapshifts$; $\overlapstaff$ specifies which staff are allowed to be assigned to $\overlapshifts$, while constraints determine if and to what degree this is respected.
For example, a staff member that is unwilling to work overlapping shifts may not be included in $\overlapstaff$. However, whether the staff member is assigned to $\overlapshifts$ or not, despite not being allowed, depends on the constraints.

The number of overlapping combinations grows rapidly with the number of overlapping shifts. Therefore, we make a simplifying assumption that if someone is allowed to be assigned to all shifts in an overlapping combination, then they are allowed to be assigned to any subset of those shifts.

\subsection{Generic Constraints}
Generic constraint (GC) formulations take input parameters of the forms $\GCparamStaff\subseteq\Staff$, $\GCparamShifts\subseteq\Shifts$, $\GCparamx, \GCparamy\in\mathbb{N}$, and $\GCparamu, \GCparamv \in \Re$.
%, determining the specific constraints that they model.
%Therefore, given specific parameter values, a GC is instanced and added to the optimization problem. 

\begin{itemize}[align=left, labelwidth=2em, leftmargin=\dimexpr\labelwidth+\labelsep\relax]
    \item[\textbf{GC1}] \emph{The number of shifts in $\GCparamShifts$ that are not assigned at least one person from $\GCparamStaff$ must be greater than or equal to $\GCparamx$ and less than or equal to $\GCparamy$.}
    \item[\textbf{GC2}] \emph{The number of shifts in $\GCparamShifts$ that have been assigned an unqualified person from $\GCparamStaff$ must be greater than or equal to $\GCparamx$ and less than or equal to $\GCparamy$.}
    \item[\textbf{GC3}] \emph{The number of times some person $\stdperson\in\GCparamStaff$ is assigned to 
    all shifts in a set of overlapping shifts $\stdoverlap\in\Overlappings$ for which they are not allowed to be assigned to ($\stdperson\not\in\overlapstaff$) must be greater than or equal to $\GCparamx$ and less than or equal to $\GCparamy$.}
    \item[\textbf{GC4}]\emph{For each person in $\GCparamStaff$, the fraction of their assigned workload from shifts in $\GCparamShifts$, if they are assigned to any at all, must be greater than or equal to $\GCparamu$ and less than or equal to $\GCparamv$.}
    \item[\textbf{GC5}] \emph{If any person in $\GCparamStaff[_1]$ is assigned to a shift in $\GCparamShifts[_1]$, then the number of assignments of people in $\GCparamStaff[_2]$ to shifts in $\GCparamShifts[_2]$ must be greater than or equal to $\GCparamx$ and less than or equal to $\GCparamy$.}
    \item[\textbf{GC6}] \emph{The number of consecutive days of shifts in $\GCparamShifts$ assigned to a person in $\GCparamStaff$ must be greater than or equal to $\GCparamx$ and less than or equal to $\GCparamy$.}
    \item[\textbf{GC7}] \emph{For a person in $\GCparamStaff$ that is assigned to shifts in $\GCparamShifts$ for $\GCparamx$ to $\GCparamy$ consecutive days, the $\GCparamn$ days immediately before must be without assignments to shifts in $\GCparamShifts[_1]$ and $\GCparamm$ days immediately after must be without assignments to shifts in $\GCparamShifts[_2]$.}
    \item[\textbf{GC8}] \emph{If a staff member in $\GCparamStaff$ is assigned to a shift in $\GCparamShifts$, then, on the day immediately before the assigned shift's start day, the staff member must either be assigned to a shift in $\GCparamShifts$ with the same shift type or not be assigned to any shift in $\GCparamShifts$.}
    % then the preceding day must either have an assignment of a shift in $\GCparamShifts$ of the same type, or no assignments of any shift in $\GCparamShifts$.}
    \item[\textbf{GC9}] \emph{For a staff member in $\GCparamStaff$, the workload that they are assigned to from shifts in $\GCparamShifts$ divided by their desired workload must be within $\GCparamv$ percent of the expected workload ratio. The expected workload ratio is the total workload of all shifts in $\GCparamShifts$ divided by the total desired workload of all staff members in $\GCparamStaff$.}
\end{itemize}
See Section~\ref{sec:ExperimentalEvaluation} for examples of constraints that these GCs can model.
We consider only hard constraints, therefore, this is a constraint satisfaction problem.

\section{Formulations}
\label{sec:Formulations}

This section presents the SMT and MILP formulations of GC1--GC3. For brevity, the remaining formulations are available at our repository\footnote{\label{GitHublinkFootnote}\GitHubLink}.
The goal is to find an assignment to all variables such that all constraints are satisfied.

% \textit{Start Day} returns the set of all shifts in a given set that start on a given day,
% \begin{equation*}
%     \ShiftFiltersd{\widetilde\Shifts, k} = \left\{ \stdshift \in \widetilde\Shifts \mid \shiftsd = k \right\}.
% \end{equation*}
% \textit{Relative Start Day} is similar to Start Day but relative to a given shift,
% \begin{equation*}
%     \ShiftFiltersdrel{\stdshift}{\widetilde\Shifts, k} = \left\{ \stdshift' \in \widetilde\Shifts \mid \shiftsd[\stdshift'] = \shiftsd + k \right\}.
% \end{equation*}
% \textit{Type Set} returns the subset of shifts with a given type,
% \begin{equation*}
%     \ShiftFiltertype{t}{\widetilde\Shifts} = 
%     \left\{ 
%         \stdshift' \in \widetilde\Shifts \mid \shifttype[\stdshift'] = t
%     \right\}
% \end{equation*}
% \textit{Relative Type Set} returns the subset of shifts of the same shift type as a given shift,
% \begin{equation*}
%     \ShiftFiltertyperel{\stdshift}{\widetilde\Shifts} = 
%     \left\{ 
%         \stdshift' \in \widetilde\Shifts \mid \shifttype[\stdshift'] = \shifttype
%     \right\}
% \end{equation*}
% Finally, $\Powerset{\Omega}=\{\Omega'\mid\Omega'\subseteq\Omega\}$ denotes the power set of $\Omega$.

\subsection{SMT Formulation}
\label{sec:SMT}
% In this SMT formulation, The SMT formulation of the GCs are described in this section. 
For each person $\stdperson\in\Staff$ and shift $\stdshift\in\Shifts$, the Boolean variable $\assignment$ denotes if $\stdperson$ is assigned to $\stdshift$. 
Every instance of a GC formulation (with specific parameter values) adds constraints to the problem. Therefore, in the following formulations, all parameters should be interpreted as specific for a GC instance. 
% The aim is to find an assignment to all variables $\assignment$, such that all constraints are satisfied.
We begin by defining useful notation.
The \textit{Boolean to Binary} operator $\boolmap$ maps $\mathit{True}\rightarrow1$ and $\mathit{False}\rightarrow0$,
\begin{equation*}
    \boolmap\,\omega = 1 \text{ if $\omega$ is true},\; 0 \text{ otherwise.}
\end{equation*}
\textit{Boolean Summation} extends this to count the number of \textit{True} evaluations for a function over a set $\Omega$,
\begin{equation*}
    \boolsum{\omega\,\in\,\Omega} f\left( \omega \right) \;=\; \sum_{\omega\,\in\,\Omega} \boolmap f\left( \omega \right).
\end{equation*}

We now formulate the constraints. Only one staff member can be assigned to a shift, formulated in~\eqref{eq:SMT:constraint:0}.
\textbf{GC1} is formulated in~\eqref{eq:SMT:constraint:1} and \textbf{GC2} in~\eqref{eq:SMT:constraint:2},
\begin{equation}
    \label{eq:SMT:constraint:0}
    \left( \boolsum{\stdperson\,\in\,\Staff}\,\assignment \right) \;\leq\; 1, \quad \forall \stdshift \in \Shifts,
\end{equation}
% \begin{equation}
%     \label{eq:SMT:constraint:0}
%     \left( \boolsum{\stdperson\,\in\,\Staff}\,\assignment \right) \leq 1, \quad \forall \stdshift \in \Shifts
% \end{equation}
\begin{equation}\label{eq:SMT:constraint:1}
    \GCparamx \;\leq\; \left( \boolsum{\stdshift \in \GCparamShifts} \boolAND{\stdperson \in \GCparamStaff}{} \neg \assignment \right) \;\leq\; \GCparamy,
\end{equation}
% \begin{equation}\label{eq:SMT:constraint:1}
%     \GCparamx \leq \left( \boolsum{\stdshift \in \GCparamShifts} \boolAND{\stdperson \in \GCparamStaff}{} \neg \assignment \right) \leq \GCparamy
% \end{equation}
\begin{equation}\label{eq:SMT:constraint:2}
    \GCparamx \;\leq\; \left( \boolsum{\stdshift \in \GCparamShifts}  \boolOR{\stdperson \in \GCparamStaff \setminus \shiftqp}{} \assignment  \right) \;\leq\; \GCparamy.
\end{equation}
% \begin{equation}\label{eq:SMT:constraint:2}
%     \GCparamx \leq \left( \boolsum{\stdshift \in \GCparamShifts}  \boolOR{\stdperson \in \GCparamStaff \setminus \shiftqp}{} \assignment \right) \leq \GCparamy
% \end{equation}

For \textbf{GC3}, we define the set $\Overlappings^{\mathit{NA}}_{\stdperson}$ containing every $\stdoverlap\in\Overlappings$ with 2-sized $\overlapshifts$ to which staff member $\stdperson$ is not allowed to be assigned,
\begin{equation}\label{eq:SMT:constraint:3.1}
    \Overlappings_{\stdperson}^{\mathit{NA}} = \left\{ \stdoverlap \in \Overlappings \,\bigg\vert\, \vert\overlapshifts\vert=2, \stdperson\not\in\overlapshifts \right\}, \quad
        \forall \stdperson \in \GCparamStaff.
\end{equation}
Thus, we constrain the sum over all staff members in $\stdperson\in\GCparamStaff$ and overlapping combinations in $\stdoverlap\in\Overlappings_{\stdperson}^{\mathit{NA}}$ where $\stdperson$ is assigned to all shifts in $\overlapshifts$:
\begin{equation}\label{eq:SMT:constraint:3.3}
    \GCparamx \;\leq\;
    \sum_{\stdperson \in \GCparamStaff} \left( \boolsum{\stdoverlap \in \Overlappings_{\stdperson}^{\mathit{NA}}} \boolAND{\stdshift \in \overlapshifts}{} \assignment \right) \;\leq\; \GCparamy.
\end{equation}

\subsection{MILP Formulation}
\label{sec:MILP}

The MILP formulation uses $0$-$1$-variables: $\assignment=1$ if person $\stdperson$ is assigned to shift $\stdshift$, $0$ otherwise. 
Besides these, additional auxiliary $0$-$1$-variables $\GCvarInter$ are used in most GC formulations, which should be interpreted as specific to each GC instance.
First, only one person can be assigned to a shift,
\begin{equation}
    \label{eq:MILP:constraint:0}
    \sum_{\stdperson \in \Staff} \assignment \;\leq\; 1, \quad \forall \stdshift \in \Shifts.
\end{equation}
\textbf{GC1}
is formulated in~\eqref{eq:MILP:constraint:1.1} and uses auxiliary variables $\GCvarInter[_\stdshift], \forall\stdshift\in\GCparamShifts$, which are modeled in~\eqref{eq:MILP:constraint:1.2} using the big-M method ($\GCconstBigM \geq \lvert \GCparamStaff \rvert$) such that $\GCvarInter[_\stdshift]=1$ if no one in $\GCparamStaff$ is assigned to shift $\stdshift$, $0$ otherwise,
\begin{equation}\label{eq:MILP:constraint:1.2}
    \begin{gathered}
        1 - \GCvarInter[_\stdshift] \;\leq\; \sum_{\stdperson \in \GCparamStaff} \assignment \;\leq\; \GCconstBigM \left( 1 - \GCvarInter[_\stdshift] \right),
        \quad \forall \stdshift \in \GCparamShifts,
    \end{gathered}
\end{equation}
\begin{equation}
    \label{eq:MILP:constraint:1.1}
    \GCparamx \;\leq\; \sum_{\stdshift \in \GCparamShifts} \GCvarInter[_\stdshift] \;\leq\; \GCparamy.
\end{equation}
\textbf{GC2}, formulated in~\eqref{eq:MILP:constraint:2.1}, uses auxiliary variables $\GCvarInter[_\stdshift], \forall\stdshift\in\GCparamShifts$, which are modeled in~\eqref{eq:MILP:constraint:2.3} using the big-M method ($\GCconstBigM \geq \lvert \GCparamStaff \rvert$) such that $\GCvarInter[_\stdshift]=1$ if at least one unqualified person in $\GCparamStaff$ is assigned to shift $\stdshift$,
\begin{equation}\label{eq:MILP:constraint:2.3}
    \GCvarInter[_\stdshift] \;\leq\; \sum_{\stdperson \in \GCparamStaff \setminus \shiftqp} \assignment \;\leq\; \GCconstBigM \GCvarInter[_\stdshift], \quad \forall \stdshift \in \GCparamShifts,
\end{equation}
\begin{equation}\label{eq:MILP:constraint:2.1}
    \GCparamx \;\leq\; \sum_{\stdshift \in \GCparamShifts} \GCvarInter[_\stdshift] \;\leq\; \GCparamy.
\end{equation}
Finally, \textbf{GC3} in~\eqref{eq:MILP:constraint:3.5} uses $\Overlappings^{\mathit{NA}}_{\stdperson}$ from~\eqref{eq:SMT:constraint:3.1}.
Auxiliary variables $\GCvarInter[_{\stdoverlap, \stdperson}]$, $\forall\stdoverlap\in\Overlappings^{\mathit{NA}}_{\stdperson}$, $\forall\stdperson\in\GCparamStaff$, are modeled in~\eqref{eq:MILP:constraint:3.3} such that $\GCvarInter[_{\stdoverlap, \stdperson}] = 1$ if $\stdperson$ is assigned to all shifts in $\overlapshifts$, $0$ otherwise,
\begin{equation}\label{eq:MILP:constraint:3.3}
    2\GCvarInter[_{\stdoverlap, \stdperson}] \;\leq\; \sum_{\stdshift \in \overlapshifts} \assignment \;\leq\; \GCvarInter[_{\stdoverlap, \stdperson}] + 1, \quad \forall \stdoverlap \in \Overlappings^{\mathit{NA}}_{\stdperson}, \forall \stdperson \in \GCparamStaff,
\end{equation}
\begin{equation}\label{eq:MILP:constraint:3.5}
    \GCparamx \;\leq\; \sum_{\stdperson \in \GCparamStaff} \sum_{\stdoverlap \in \Overlappings^{\mathit{NA}}_{\stdperson}} \GCvarInter[_{\stdoverlap, \stdperson}]
    \;\leq\; \GCparamy.
\end{equation}

\section{Experimental Evaluation}
\label{sec:ExperimentalEvaluation}
We compare the SMT and MILP formulations using the solvers Z3 (version 4.8.12.0) and Gurobi (version 9.1.2), respectively, on two different types of rostering problems. As Gurobi uses multi-core processing by default, this was enabled for Z3. Apart from that, both solvers use default settings without customizations. 
All experiments are run on a 2020 MacBook Air, 8-core Apple M1, 16 GB RAM, macOS Monterey 12.0.1. See our repository for additional details\footref{GitHublinkFootnote}.

\subsection{Problem A}

Problem~A is inspired by~\cite{santos2016integer} and~\cite{burke2008hybrid}, which detail typical NRPs. 
Instances are generated from a given number of shifts and staff members.
Shifts from Table~\ref{tab:exp:problemA_shifts} are added sequentially, from top to bottom, to the first day. Once the end of the table is reached, the same is done for the next day, and so on, until the desired total number of shifts is achieved.
Similarly, staff members from Table~\ref{tab:exp:problemA_staff} are added sequentially, cycling through the list repeatedly, until the problem contains the desired number of staff members.

This problem includes six daily shifts: $2$ day, $2$ evening, and $2$ night shifts, each lasting $9$ hours with equal workload. Shifts overlap with adjacent shifts. All shifts require the \textit{Nurse} qualification (N), while one shift also requires \textit{Administration} (A). In Table~\ref{tab:exp:problemA_staff}, one person has both qualifications and another has half the desired workload. Overlapping shifts are disallowed for all personnel. The constraints for Problem~A are detailed in Table~\ref{tab:problemA_constraints}, along with the GC and input parameters used to model them.
\begin{table}[]
    \centering
    \caption{Set of shifts for problem A, with each respective type, start hour, duration, workload (WL) in hours, and required qualifications.}
    \label{tab:exp:problemA_shifts}
    \begin{tabular}{@{}lcccl@{}}
        \toprule
        Type & Start & Dur. & WL & Req.Qual. \\
        \midrule
        \ProblemShiftNameNurseD{1} & 06:00 & 9:00 & 9 & N, A \\
        \ProblemShiftNameNurseD{2} & 06:00 & 9:00 & 9 & N \\
        \ProblemShiftNameNurseE{1} & 14:00 & 9:00 & 9 & N \\
        \ProblemShiftNameNurseE{2} & 14:00 & 9:00 & 9 & N \\
        \ProblemShiftNameNurseN{1} & 22:00 & 9:00 & 9 & N \\
        \ProblemShiftNameNurseN{2} & 22:00 & 9:00 & 9 & N \\
        \bottomrule
    \end{tabular}
\end{table}
\begin{table}[]
    \centering
    \caption{Set of staff member types for problem A, with each respective desired workload in hours, and qualifications.}
    \label{tab:exp:problemA_staff}
    \begin{tabular}{@{}cll@{}}
        \toprule
        Type & Desired WL & Qualifications \\
        \midrule
        1 & 100 & N, A \\
        2 & 100 & N \\
        3 & 100 & N \\
        4 & 50  & N \\
        \bottomrule
    \end{tabular}
\end{table}

\begin{table}[]
    \small
    \renewcommand{\arraystretch}{1.2}
    \setlength{\tabcolsep}{3pt}
    \caption{Constraints for problem A.}
    \label{tab:problemA_constraints}
    \begin{tabular}{@{}>{\arraybackslash}p{0.05\columnwidth}@{} >{\arraybackslash}p{0.95\columnwidth}@{}}
      \toprule
      1. & All shifts must be assigned at least one person; \textbf{GC1}:  $\GCparamShifts = \Shifts$, $\GCparamStaff = \Staff$, $\GCparamx=\GCparamy=0$.\\
      2. & The $1^\text{st}$ staff member cannot be assigned to any shift that starts on days 3--5 of the schedule; \textbf{GC1}: $\GCparamShifts = \ShiftFiltersd{\Shifts, \{3,4,5\}}$, $\GCparamStaff = \{1\}$, $\GCparamx=\GCparamy=\lvert \ShiftFiltersd{\Shifts, \{3,4,5\}} \rvert$.\\ 
      3. & Everyone must be qualified for the shifts that they are assigned to; \textbf{GC2}: $\GCparamShifts = \Shifts$, $\GCparamStaff = \Staff$, $\GCparamx=\GCparamy=0$. \\
      % 4. & No one should be assigned to overlapping shifts that they are not allowed to be assigned to simultaneously; \textbf{GC3}. \\
      4. & No one should be assigned to shifts that overlap in time unless they are allowed to; \textbf{GC3}: $\GCparamStaff = \Staff$, $\GCparamx=\GCparamy=0$.
      \\
      5. & At least 50\% of the shifts assigned to the $1^\text{st}$ staff member must be of the types \emph{\ProblemShiftNameNurseD{1}} and \emph{\ProblemShiftNameNurseD{2}}; \textbf{GC4}: $\GCparamShifts = \ShiftFiltertype{\{\textit{\ProblemShiftNameNurseD{1}, \ProblemShiftNameNurseD{2}}\}}{\Shifts}$, $\GCparamStaff = \{1\}$, $\GCparamu=0.5$, $\GCparamv= 1$. \\
      6. & If the $1^\text{st}$ staff member is assigned to any shift starting on days 1 or 2, the $4^\text{th}$ and $7^\text{th}$ staff members may not be assigned to any shifts on those days; \textbf{GC5}: $\GCparamShifts[_1] = \GCparamShifts[_2] = \ShiftFiltersd{\Shifts, \{1, 2\}}$, $\GCparamStaff[_1] = \{0\}$, $\GCparamStaff[_2] = \{4, 5\}$, $\GCparamx=\GCparamy=0$. \\
      7. & If someone has been assigned a day shift, they cannot be assigned a night shift on the same day; $\forall \stdperson \in \Staff$, $\forall \GCvard \in \Startdays$: \textbf{GC5}: $\GCparamShifts[_1] = \ShiftFiltersd{\Shifts, \GCvard}  \cap \ShiftFiltertype{\{\textit{\ProblemShiftNameNurseD{1}, \ProblemShiftNameNurseD{2}}\}}{\Shifts}$, $ \GCparamShifts[_2] = \ShiftFiltersd{\Shifts, \GCvard}  \cap \ShiftFiltertype{\{\textit{\ProblemShiftNameNurseN{2}, \ProblemShiftNameNurseN{2}}\}}{\Shifts}$, $\GCparamStaff[_1] = \GCparamStaff[_2] = \{\stdperson\}$, $\GCparamx=\GCparamy=0$. \\
      8. & No one can be assigned to shifts for more than 6 days in a row; \textbf{GC6}: $\GCparamShifts = \Shifts$, $\GCparamStaff = \Staff$, $\GCparamx=0$, $\GCparamy=6$. \\
      9. & If someone has been assigned to shifts for 4--6 days in a row, they must be off 3 days before and after; \textbf{GC7}: $\GCparamShifts = \GCparamShifts[_1] = \GCparamShifts[_2] = \Shifts$, $\GCparamStaff = \Staff$, $\GCparamx=4$, $\GCparamy=6$, $\GCparamn=\GCparamm=3$. \\
      10. & All consecutive days of shift assignments must be of the same type; \textbf{GC8}: $\GCparamShifts = \Shifts$, $\GCparamStaff = \Staff$. \\
      11. & All personnel workloads must be within 30\% of the expected workload, with regards to all shifts; \textbf{GC9}: $\GCparamShifts = \Shifts$, $\GCparamStaff = \Staff$, $\GCparamv=0.3$. \\
      \bottomrule
    \end{tabular}
\end{table}

\begin{figure}[]
    \centering
    \includegraphics[width=0.47\textwidth]{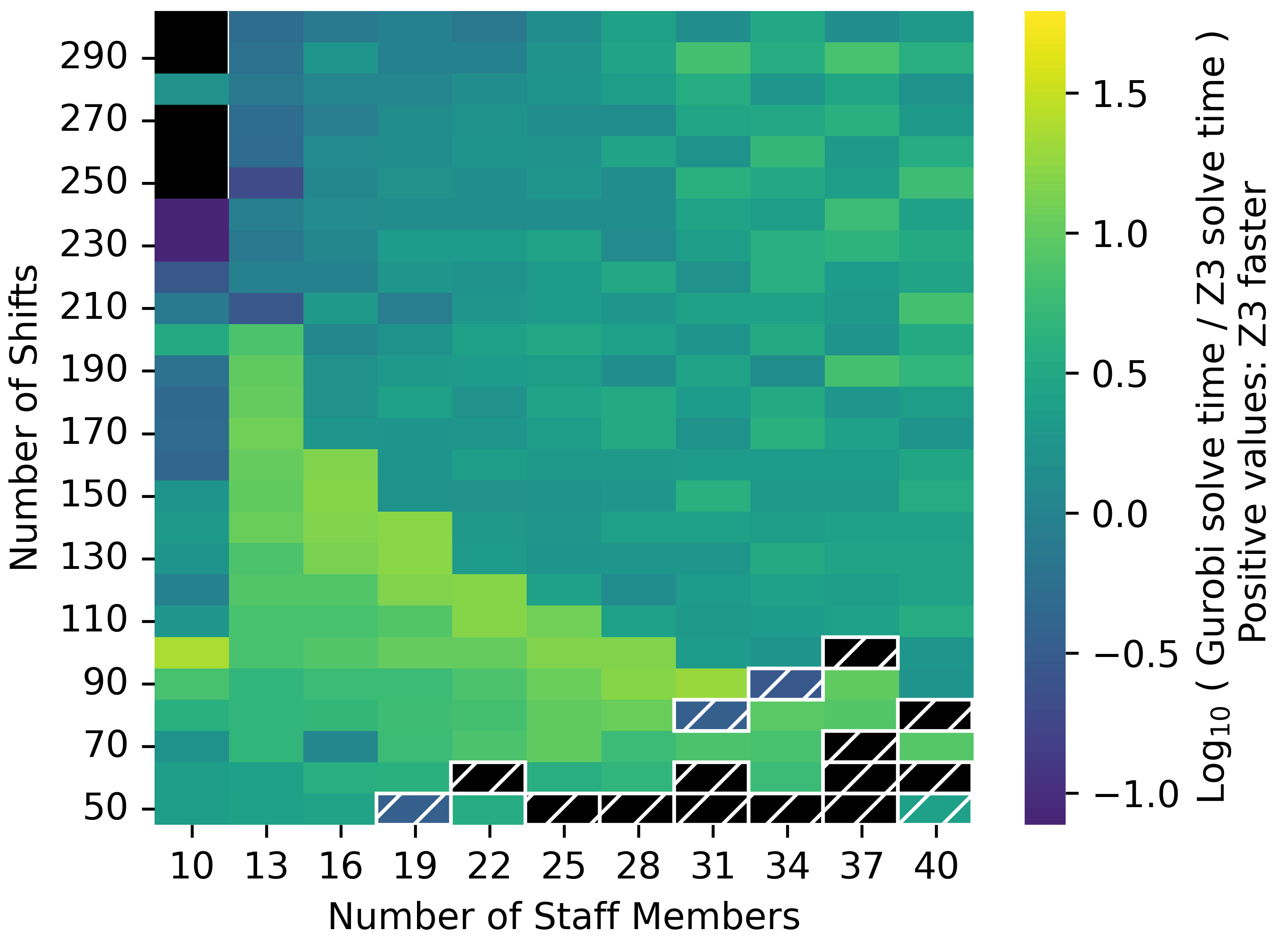}
    \caption{A comparison of the solving times between Z3 and Gurobi for each problem instance. Black means Z3 was not able to find a solution within the time limit while Gurobi could. White hatching means the problem is infeasible.}
    \label{fig:ProblemA_Comparison}
\end{figure}

Instances of the problem are created for the number of shifts ranging from $50$--$300$, equivalent to approximately $8$--$50$ scheduled days, and the number of staff members from $10$--$40$. 
The solvers are given a time limit of 1 hour to either find a solution or determine that the problem is infeasible. If they fail to return a result within this time frame, the instance is considered unsolved.
%The solvers have a time limit of $1$ hour to terminate with a solution or infeasibility.

\begin{figure}[h]
    \centering
    \includegraphics[width=0.47\textwidth]{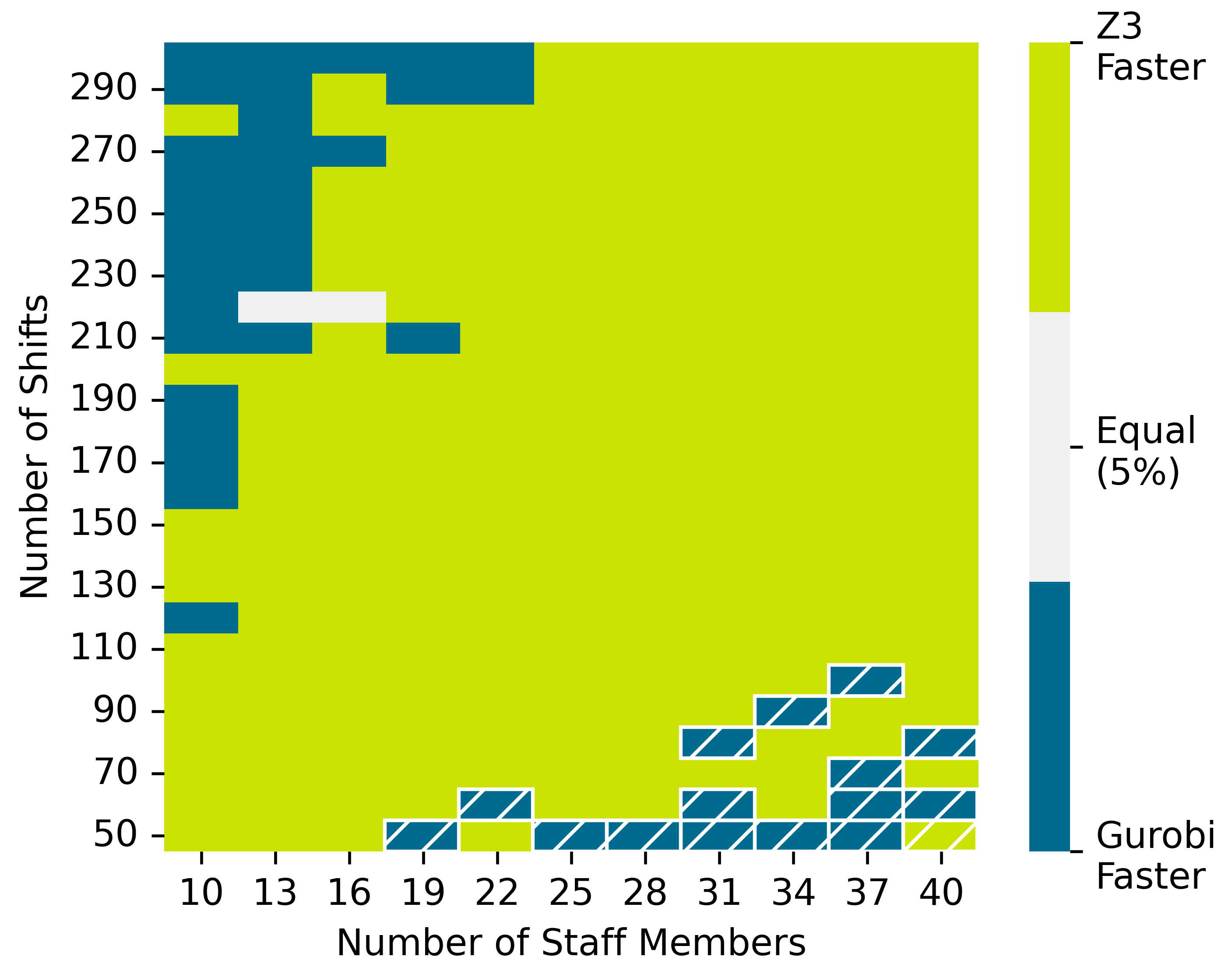}
    \caption{A comparison between Z3 and Gurobi based on solving times for each problem instance. White hatching means the instance is infeasible.}
    \label{fig:ProblemA_Ranking}
\end{figure}

Figure~\ref{fig:ProblemA_Comparison} compares solver performance using the logarithmic quotient $\log_{10}(\text{Gurobi time} \,/\, \text{Z3 time})$ across varying numbers of staff members and shifts. Positive values indicate Z3 is faster, while negative values indicate Gurobi is faster. 
Black regions represent cases where Z3 failed to terminate within the time limit but Gurobi succeeded. For no instances did Z3 terminate within the time limit but not Gurobi.
White hatching denotes infeasible instances. Figure~\ref{fig:ProblemA_Ranking} shows the same instance matrix but clarifies which solver is faster (or equal within $5$\%).

The results reveal distinct performance patterns: Gurobi outperforms Z3 in regions with fewer staff members relative to shifts (top left) and fewer shifts relative to staff members (bottom right). 
It was found that relaxing constraint $11$, which bounds assigned workloads to near desired workloads, made all infeasible problems feasible and lead to both solvers terminating within the time limit. 
Thus, many shifts relative to staff members puts assigned workloads closer to their upper bounds while few shifts relative to staff members puts assigned workloads closer to their lower bounds. 
This suggests that Gurobi is faster to process the instances which are more tightly constrained by constraint $11$.
In the remaining areas, such as near the center where the problems are not as constrained, Z3 generally finds solutions faster than Gurobi.

\subsection{Problem B}
\label{sec:ProblemB}
Problem~B is larger and attempts to mirror a more complex scheduling problem in a hospital ward. Inspiration for the constraints come from the real-world case presented in~\cite{combrink_do_2021}. 
The staff set is constant, containing $29$ staff members with a mix of qualifications and desired workloads (Table~\ref{tab:exp:problemB_staff}).
The number of desired days in the schedule decides the set of shifts; the top $19$ shifts in Table~\ref{tab:exp:problemB_shifts} are added to each day in the schedule. An additional \emph{\ProblemShiftNameAdmin} shift occurs every second day, starting on the first day. 
\begin{table}[]
    \centering
    \caption{Set of staff members for problem B, with each respective desired workload in hours, and qualifications.}
        \begin{tabular}{lcl}
            \toprule
             ID & Desired WL & Qualifications \\ \midrule
            1-3. & 100 & N, CN \\
            4. & 100 & N, A \\
            5-14. & 100 & N \\
            15. & 75 & N \\
            16-17. & 50 & N \\
            18. & 100 & D, S1, S2 \\
            19-20. & 100 & D, S1 \\
            21-22. & 100 & D, S2 \\
            23. & 75 & D, S2 \\
            24-26. & 100 & D \\
            27. & 50 & D \\
            28. & 100 & A \\
            \bottomrule
        \end{tabular}
    \label{tab:exp:problemB_staff}
\end{table}
\begin{table}[]
    \centering
    \caption{Set of shifts for problem B, with each respective type, start hour, duration, workload (WL) in hours, and required qualifications.}
    \begin{tabular}{llcccl}
        \toprule
        \textbf{} & Type & Start & Dur. & WL & Req.Qual. \\ \midrule
        1-5. & \ProblemShiftNameNurseD & 06:00 & 9:00 & 9 & N\\
        6-9. & \ProblemShiftNameNurseE & 14:00 & 9:00 & 9 & N\\
        10-11. & \ProblemShiftNameNurseN & 22:00 & 9:00 & 7 & N\\
        12-13. & \ProblemShiftNameDoctorD & 06:00 & 10:00 & 10 & D\\
        14. & \ProblemShiftNameDoctorE & 12:00 & 10:00 & 10 & D\\
        15. & \ProblemShiftNameDoctorN & 19:00 & 9:00 & 7 & D\\
        16. & \ProblemShiftNameDoctorS{1} & 05:00 & 9:00 & 9 & D, S1\\
        17. & \ProblemShiftNameDoctorS{2} & 05:00 & 9:00 & 9 & D, S2\\
        18. & \ProblemShiftNameCNurseD & 06:00 & 10:00 & 10 & CN\\
        19. & \ProblemShiftNameCNurseE & 16:00 & 8:00 & 8 & CN\\
        20. & \ProblemShiftNameAdmin * & 08:00 & 5:00 & 5 & A\\
        \bottomrule
    \end{tabular}
    \\
    * Occurs every second day, starting on the first day.
    \label{tab:exp:problemB_shifts}
\end{table}

The constraints are given in Table \ref{tab:problemB_constraints}. Parameter values are deferred to our repository\footref{GitHublinkFootnote}.
\emph{\ProblemShiftNameNurseD}-\emph{\ProblemShiftNameCNurseD},
Staff members are allowed to be assigned to any of the following shift type pairs simultaneously: 
\emph{\ProblemShiftNameNurseE}-\emph{\ProblemShiftNameCNurseE},
\emph{\ProblemShiftNameDoctorD}-\emph{\ProblemShiftNameDoctorS{1}},  
\emph{\ProblemShiftNameDoctorD}-\emph{\ProblemShiftNameDoctorS{2}},
and
\emph{\ProblemShiftNameCNurseD}-\emph{\ProblemShiftNameAdmin}.

\begin{table}[]
    \caption{Constraints for problem B.}
    \begin{tabular}{@{}>{\arraybackslash}p{0.07\columnwidth}@{} >{\arraybackslash}p{0.93\columnwidth}@{}}
      \toprule
      1. & All shifts must be assigned at least one person; \textbf{GC1}.\\
      2. & Everyone must be qualified for the shifts that they are assigned to; \textbf{GC2}. \\
      3. & No one should be assigned to overlapping shifts that they are not allowed to work at the same time; \textbf{GC3}. \\
      4. & At least 10\% of the shifts assigned to person 1 (1st staff cycle) must be of the types \emph{\ProblemShiftNameNurseD} and \emph{\ProblemShiftNameNurseE}; \textbf{GC4}.\\
      5. & If someone is assigned to a \emph{\ProblemShiftNameDoctorD}, \emph{\ProblemShiftNameDoctorS{1}} or \emph{\ProblemShiftNameDoctorS{2}} shift, they cannot be assigned to a \emph{\ProblemShiftNameDoctorN} shift on the same day; \textbf{GC5}.\\
      6. & If someone is assigned to a \emph{\ProblemShiftNameNurseD} or \emph{\ProblemShiftNameCNurseD} shift, they cannot be assigned to a \emph{\ProblemShiftNameNurseN} or \emph{\ProblemShiftNameCNurseE} shift on the same day; \textbf{GC5}.\\
      7. & If someone is assigned to an \emph{\ProblemShiftNameAdmin} shift, they cannot be assigned to a \emph{\ProblemShiftNameNurseE}, \emph{\ProblemShiftNameNurseN}, \emph{\ProblemShiftNameCNurseE} or \emph{\ProblemShiftNameDoctorN} shift on the same day or on the day before; \textbf{GC5}.\\
      8. & No one can be assigned to shifts for more than 6 days in a row; \textbf{GC6}. \\
      9. & No one can be assigned to night/evening shifts for more than 3 days in a row; \textbf{GC6}. \\
      10. & If someone has been assigned to shifts for 3-5 days in a row, they must be off 2 days before and after; \textbf{GC7}. \\
      11. & If someone has been assigned to shifts for 6 days in a row, they must be off 3 days before and after; \textbf{GC7}. \\
      12. & All consecutive days of shift assignments, excluding \emph{\ProblemShiftNameAdmin} shifts, must be of the same type; \textbf{GC8}. \\
      13. & All workloads for nurses must be within 60\% of the expected workload, with respect to to all nurse shifts; \textbf{GC9}. \\
      14. & All workloads for doctors must be within 60\% of the expected workload, with respect to to all doctor shifts; \textbf{GC9}.\\
      \bottomrule
    \end{tabular}
    \label{tab:problemB_constraints}
\end{table}

\begin{figure}[h]
    \centering
    \includegraphics[width=0.49\textwidth]{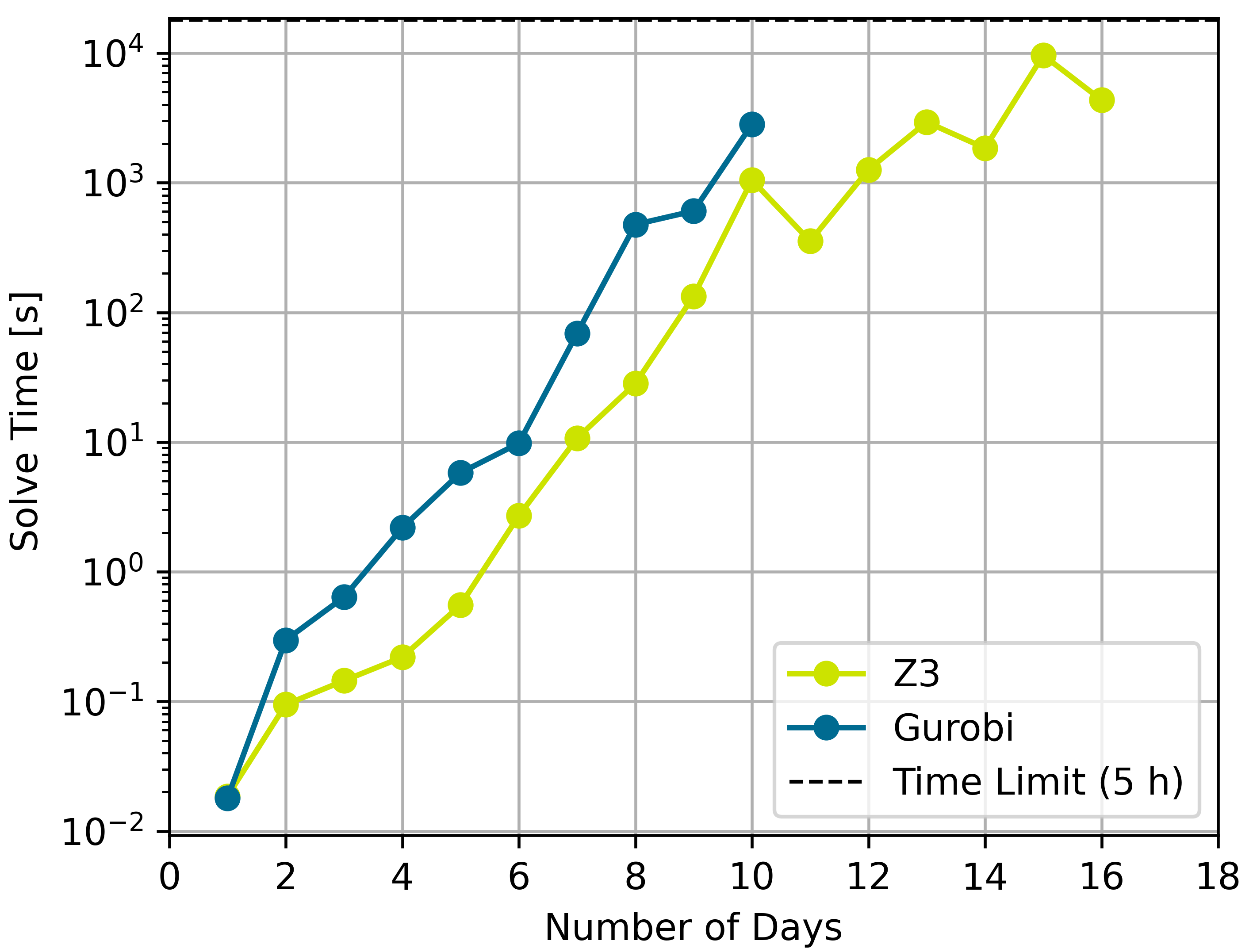}
    \caption{Solving times for both Z3 and Gurobi over the number of days included in the schedule for Problem~B. The number of days ranges from 1--17. If no data point is shown then the time limit of 5 hours was reached.}
    \label{fig:ProblemB_Times}
\end{figure}

Figure~\ref{fig:ProblemB_Times} shows the solving times for each solver over $1$--$17$ days included in the scheduling problem, with an arbitrary time limit of $5$ hours. Z3 is able to find a solution faster than Gurobi in all instances. None of the solvers were able to find solutions to problems including more than $16$ days within the time limit.

% \section{Discussion}
% \label{sec:Discussions}
% \input{Discussions}

\section{Discussions \& Conclusions}
\label{sec:Conclusions}

The GCs were formulated to maximize each respective solver's performance. In this regard, Z3 showed more sensitivity than Gurobi. 
% While implementing the constraints, 
% Z3 showed more sensitivity to how the GCs where formulated compared to Gurobi. 
% Z3's performance demonstrated significantly more sensitivity to model formulation compared to Gurobi. 
Integer-only formulations for GC4 and GC9 (achieved by scaling and rounding) lead to significant improvements on solving times for Z3 while negligibly affecting Gurobi. 
This suggests that Z3 leverages efficient integer-based solution strategies when available. Moreover, Z3 showed marked sensitivity to implementation details, with explicit arithmetic expressions sometimes outperforming built-in functions despite their high-level equivalence. These performance variations highlight that, when using Z3, careful consideration of the constraint implementation is essential. For practitioners, this implies that some experimentation with alternative formulations may be necessary to achieve better performance, particularly for complex scheduling problems like those encountered in healthcare environments.

Although this work focuses exclusively on hard constraints, real-world schedules rely heavily on soft constraints. As a next step, adopting the approach of~\cite{combrink_do_2021} by weighting GCs would introduce soft constraints while leveraging their expressiveness.
Furthermore, hybrid approaches which take advantage of in-domain knowledge, heuristics, local search techniques, etc., offer several promising directions toward improved performance and scalability.

    To conclude, this article presents a set of generic constraints that can model a wide range of rostering constraints. 
    Furthermore, SMT and MILP formulations are described and used to model two sets of problems, one based on those found in the literature and one based on a real-world setting. 
    A comparison of the state-of-the-art SMT and MILP solvers Z3 and Gurobi, respectively, is then done on the problem sets. 
    Results show how each solver's performance depends on the problem; Gurobi appears to excel on infeasible or highly constrained problems, while Z3 performs better on more relaxed problems. 
    Therefore, the specific problem to solve determines which solver and underlying framework (SMT/MILP) is more suitable. 
    Connecting back to this article's original aim of exploring SMT for the nurse rostering problem, we argue that SMT offers a promising direction for future research.
    Z3's faster solving times on the real-world based problem~B of Section~\ref{sec:ProblemB} suggest that SMT-based methods could provide value to real-world healthcare service providers, and further the work toward more sustainable and efficient healthcare.

%%%%%%%%%%%%%%%%%%%%%%%%%%%%%%%%%%%%%%%%%%%%%%%%%%%%%%%%%%%%%%
\balance
\printbibliography

\end{document}